\documentclass[openacc]{rstransa}

\usepackage{duckuments}
\usepackage{algorithm}
\usepackage{algpseudocode}

\titlehead{Preprint}

\begin{document}

\title{Peano: Learning Formal Mathematical Reasoning}

\author{
Gabriel Poesia$^{1}$ and Noah D. Goodman$^{12}$}

\address{$^{1}$Department of Computer Science, Stanford University\\
$^{2}$Department of Psychology, Stanford University}



\corres{Gabriel Poesia\\
\email{poesia@cs.stanford.edu}}



\begin{abstract}

General mathematical reasoning is computationally undecidable,
but humans routinely solve new problems.
Moreover, discoveries developed over centuries
are taught to subsequent generations quickly.
What structure enables this, and how might that inform
automated mathematical reasoning?
We posit that central to both puzzles
is the structure of procedural abstractions underlying mathematics.
We explore this idea in a case study on 5 sections of beginning algebra
on the Khan Academy platform.
To define a computational foundation,
we introduce Peano, a theorem-proving environment
where the set of valid actions at any point is finite.
We use Peano to formalize introductory algebra problems and axioms,
obtaining well-defined search problems.
We observe existing reinforcement learning methods for symbolic reasoning
to be insufficient to solve harder problems.
Adding the ability to induce reusable abstractions (``tactics'')
from its own solutions allows an agent to make steady progress,
solving all problems.
Furthermore, these abstractions induce an order to the problems,
seen at random during training.
The recovered order has significant agreement with the expert-designed 
Khan Academy curriculum, and second-generation agents trained on the
recovered curriculum learn significantly faster.
These results illustrate the synergistic role of abstractions and
curricula in the cultural transmission of mathematics.

\end{abstract}

\begin{fmtext}

\end{fmtext}

\maketitle

\section{Introduction}

A wide range of important human questions are formulated in the language of mathematics: problems from physics to economics to computation.
Thus, computers that are able to reason about mathematical problems could have broad impact
on scientific questions.
Such systems could be used directly by scientists or could potentially aid in training the next generation of human scientists and engineers.
Existing systems, such as Computer Algebra Systems, implement procedures for solving  important classes of well-known mathematical problems, such as differential equations or matrix
computations. A much more ambitious goal would be to have computers
that can reason about novel types of problems. How could that be done?

There are several computational foundations for expressing mathematics that can serve as a basis for
this endeavour, such as type systems and first-order or higher-order logics.
Given any of these formal systems, we can make new definitions, state propositions
and specify rules of inference.
Together, these components pose a well-defined computational problem:
given a mathematical proposition $p$, determine whether it is provable under the given 
assumptions.
Unfortunately, no algorithm can decide this for general mathematical propositions:
this is precisely Hilbert's \emph{Entscheidungsproblem},
answered in the negative by both Church and Turing even before modern computers had been invented.

In spite of the impossibility of a general procedure, humans routinely find
solutions to novel mathematical problems --- either new to them or to all mathematicians.
Moreover, new generations of mathematicians can reach the frontiers of mathematical
knowledge notably faster than the time it took previous generations to discover them.
These puzzles suggest some structure to human mathematics that makes new
problems approachable and their solutions teachable,
even if we might not hope to solve arbitrary mathematical problems.

To investigate this structure,
we propose a case study on learning to solve
educational mathematical problems given minimal prior knowledge.
More concretely, we set to formalize 5 sections of the algebra curriculum
available on the Khan Academy educational platform.
We then aim to automatically solve exercises from those sections in a system
with no prior knowledge on \emph{how} to find solutions, aside from the
basic axioms from which solutions can be formed.
Focusing on this educational domain serves two purposes: It allows us to attack the general problem of mathematical reasoning from a starting point that must be solvable -- even a child can do it! And, it means that successes at learning to generate human-like solutions have immediate applications to education.

Many modern theorem-proving languages, such as Lean, Coq or Isabelle,
could be use to \emph{represent} the axioms and problems we aim to study.
However, current languages pose a challenge in setting up
the problem of \emph{searching} for solutions:
they permit an infinite number of valid proof steps at any given point.
To conduct our case study, we start by proposing a theorem-proving
language, called Peano, designed with a companion environment for general proof search.
This environment exposes a space of valid proof steps that is finite at all times,
as we describe in detail in Section~\ref{sec:peano}.
Peano uses dependent types to encode mathematical definitions
and proofs in a general fashion using the paradigm of
\emph{propositions as types, proofs as programs}.
Using Peano, we formalize 5 sections of Khan Academy --- their axioms and exercises ---
obtaining a family of well-defined search problems.

How might we solve these problems in a computer agent?
Unguided search can only find the most trivial solutions because the search space grows rapidly with solution depth,
as for most formal systems.
Even short solutions, however, reveal patterns about when axioms are useful,
suggesting we can attempt to learn from past searches to improve our chances of solving
harder problems.
More specifically, we can guide search using a \emph{policy}: a distribution over actions to take given
the current state.
A policy can be used to prioritize taking actions that are more likely to lead to a solution.
Learning a policy from experience is the central problem of \emph{reinforcement learning}. We can thus use techniques from reinforcement learning to train an agent to solve problems in our mathematical domain.
Indeed, we observe in section \ref{sec:learning} that an agent trained using an existing reinforcement learning
method can learn to solve all problems in the first two Khan Academy sections.
However, it fails to make progress on harder problems beyond the first sections.
Indeed, as we look at later problems in the Khan Academy curriculum, axiomatic solutions
become steadily longer and thus progressively less likely to be found by exploration.
This problem would grow worse were we to continue on the human curriculum,
even if we restricted ourselves to later sections that do not require additional axioms.

This difficulty, however, is certainly not a barrier for human students, who
routinely learn this material and much beyond.
What structure in these problems might they use that an agent could leverage to make progress?
We show that adding the ability to abstract patterns in previously found
solution into new atomic, higher-level actions makes the problem tractable.
We present a simple algorithm based on anti-unification in Section~\ref{sec:tactic-induction}
for learning \emph{tactics} --- higher-level solution actions that can invoke axioms or other tactics.
Given actions at the right level of abstraction, all problems we study admit short formal solutions,
which can thus be found within a reasonable search budget.
When attempting combinations of existing tactics, success at new problems helps reveal
which new combinations are useful, further suggesting newer tactics.
This process allows the agent to make progress without being given any additional
information about the problems themselves.

Furthermore, we posit that the tactics our agents learn reflect some of the ordering
given in the Khan Academy curriculum,
\emph{despite that ordering not having been given to the agent during training}.
When we re-order problems based on dependencies between the learned tactics that appear in their solutions,
as we describe in Section~\ref{sec:curriculum},
we find that the recovered order agrees with the order of problems on Khan Academy to a significant level.
We finally observe a virtuous interaction between tactic learning and curricula.
When training a second agent again using tactic learning, but this time seeing problems as ordered
by the curriculum constructed from the first agent, we find that this second agent learns
significantly faster.
This experiment paints a computational account of the role of curricula in cultural transmission
of mathematics: abstractions that might take long to be developed can be efficiently taught
when the right ordering of educational experiences is put together.

In summary, we make the following contributions:
\begin{itemize}
    \item We introduce Peano, a theorem-proving language based on dependent types
    along with an environment for proof search where the action space is finite.
    \item We formalize 5 sections of the Khan Academy algebra curriculum in Peano,
    and show that the resulting search problems are challenging for a reinforcement learning agent.
    \item We show that a \emph{tactic induction} algorithm, where patterns in previous successful
    searches give rise to new atomic actions, enables an agent to make steady progress through
    the Khan Academy problems, eventually learning to solve them all.
    \item We observe that the tactics our agent learns can be used to largely reconstruct the order of problems in Khan Academy, and that training an agent on the resulting
    order enables faster learning.
\end{itemize}

\section{Related Work}

Automated theorem proving was one of the first targets of artificial intelligence
research. The Logic Theorist, a program developed by Newel, Simon and Shaw  \cite{mccorduck2004machines,russell2010artificial},
was demonstrated at the Darthmouth Summer Research Project on Artificial Intelligence in 1956 \cite{mccarthy2006proposal}, the meeting that first coined the name ``Artificial Intelligence'' for the
field. This work pioneered the idea of formulating mathematical reasoning problems as
search in a state graph, where edges correspond to possible deductions.
These ideas were later extended to other richer formal systems, such as
first-order logic and Euclidean geometry \cite{newell1959report}.
These programs quickly surfaced the need for search heuristics: even in a simple
domain such as propositional logic, unguided search algorithms can only find
solutions to trivial problems because of the combinatorial explosion of the search space.
While heuristics can push this limit, the need to manually engineer problem-solving
strategies for each problem domain hindered progress and interest on this research program
\cite{robinson1963theorem}.

These initial efforts on automated reasoning aimed to create programs
that solved problems in general domains by making sequences of human-like deduction steps.
Two deviations from this paradigm led to progress in different directions.
First, focusing on constrained problems
--- such as satisfiability (SAT), in propositional logic or other theories ---
led to algorithms such as DP \cite{davis1960computing} and DPLL \cite{davis1962machine}.
While even SAT is NP-Complete, these methods and later developments were able to
solve increasingly large practical instances, enabling a range of applications
in areas such as model checking and program verification.

Towards using computers in general mathematics, full automation of reasoning
--- such as what SAT and SMT solvers aim to provide --- has proved much more difficult \cite{robinson1963theorem}.
But significant progress has been made in the development of interactive proof assistants,
where humans guide the proof generation process with potential aid for completing lower level
details\footnote{In fact, in modern proof assistants such as Isabelle/HOL, SMT solvers
can be called to find proofs for sub-problems that are expressible in a theory they support.}.
Modern proof assistants --- such as Coq, Lean, Isabelle and HOL Light --- have enabled the
formalization of large bodies of
complex mathematical results; examples include a formal proof of the Kepler conjecture
\cite{hales2017formal} in the HOL Light and Isabelle proof assistants,
and a proof of the independence of the continuum hypothesis \cite{han2020formal}
in Lean.
Thus, their logical foundations have been shown to express and verify
complex mathematical results.
Even if their power also makes automation difficult in general, these languages
allow users to define \emph{tactics}: programs that can encode complex proof strategies,
and thus implement domain-specific automation to assist proving theorems in the target domain.

Concurrently, machine learning methods have advanced substantially over the last decades,
both in supervised learning \cite{krizhevsky2017imagenet},
where a model is trained to fit a given dataset,
and in reinforcement learning (RL),
where the learner's goal is choose actions that maximize reward signals
in a dynamic environment \cite{mnih2013playing,anthony2017thinking,silver2018general}.
In particular, deep neural networks trained via RL emerged as a tool to learn
policies (distributions over actions to take given a state) and value functions
(an estimate of rewards that can be obtained from a given state) from raw representations,
such as strings or pixels.
Both policies and value functions can be used to guide search algorithms,
making deep RL suitable for learning to search in large-scale problems
such as finding proofs \cite{whalen2016holophrasm,bansal19holist, huang2018gamepad}.
Given the availability of proofs generated by human mathematicians in large formalization projects,
researchers have explored the use of human-written proofs as supervised training data to
guide proof search \cite{irving2016deepmath,whalen2016holophrasm,bansal19holist}.
There has been little experience, however, in having mathematicians use these systems
outside of the domains they have been trained on,
as is the case in new formalization projects.
To eliminate or mitigate this dependency on pre-existing training data,
research efforts have also focused on the application of deep reinforcement learning
to theorem proving, as a means to learn from experience by interacting with an environment
\cite{kaliszyk2018reinforcement,wu2021tacticzero,polu2022formal}.
Systems that learn without prior training data, however, have only been demonstrated
in more constrained formal systems, such as 
first-order logic \cite{kaliszyk2018reinforcement} or a selection of a few
tactics in HOL4 \cite{wu2021tacticzero}.

Finally, another related thread of research that has seen significant progress
recently is the field of program synthesis, where the goal is to generate computer programs
that satisfy a given specification.
Programs can be mapped to mathematical proofs
by encoding of mathematical propositions as types, and proofs as programs.
This is known as the Curry-Howard correspondence,
which we explore in more depth in Section~\ref{sec:peano}.
An important insight of recent program synthesis systems, such as DreamCoder \cite{ellis2021dreamcoder},
is that learning to search for programs can be interspersed with
\emph{library learning}, whereby the synthesizer extracts useful reusable patterns
--- abstractions --- from programs that it managed to synthesize.
When solving a family of related problems, discovering abstractions
can drastically reduce the difficulty of harder problems: even complex
programs can have short implementations in terms of the appropriate library.
This insight has a direct interpretation in the land of mathematical reasoning:
the difficulty of a mathematical problem also heavily depends on the ``library''
--- of existing lemmas or tactics. Given any starting library, harder problems
can require arbitrarily large search depths to be solved.
But if we have a family of related problems of progressive difficulty,
learning abstractions --- such as the tactics we induce in Section~\ref{sec:tactic-induction} ---
is a means to make steady progress.

\section{Overview}

\begin{figure}
    \centering
    \includegraphics[width=\textwidth]{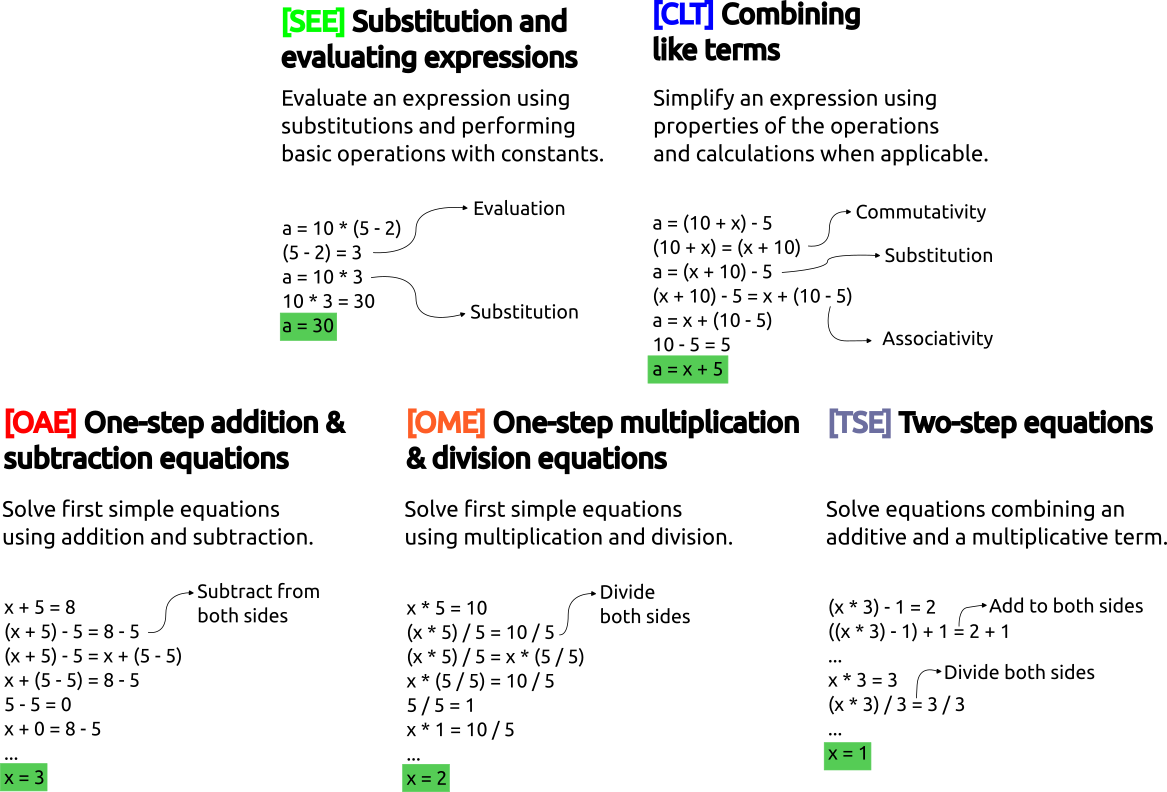}
    \caption{Illustration of the formalized versions of five sections of the
    algebra curriculum on the Khan Academy educational platform.}
    \label{fig:ka-sections}
\end{figure}

Consider the five sections from Khan Academy shown in Figure~\ref{fig:ka-sections}.
These sections take a student that knows the basic operations with numbers
--- addition, subtraction, multiplication, division --- as well as their basic properties --- commutativity, associativity, identity elements, and so on --- and teaches them to solving their first equations,
such as $x + 5 = 8$. We aim to solve these problems in a formal system where
each deduction corresponds to a complete solution step, akin to a line
that a student could write on paper.
We start this endeavour with two goals in mind.
First, this process might give insight into what ingredients a learning system would need
to do human mathematics (since general mathematical reasoning is undecidable but human students learn with alacrity).
Second, this system can then be used as a foundation for helping students
that routinely need to go through these problems --- by checking their work, providing hints, and
creating worked examples, among other pedagogical actions.

To start, we need a computational basis for representing problems and solutions.
Modern theorem-proving languages are a natural option:
their logical foundations are powerful enough
to express mathematics even at the research frontier,
from definitions to propositions and proofs.
However, when seen as environments for automatically finding solutions, these languages
pose several challenges. First, given a solution in progress, there's often an infinite
number of next steps that are valid (we dissect the reasons for this in Section~\ref{sec:peano}).
This fact implies that agents cannot learn by choosing actions from a list of valid options
-- they need to \emph{generate} their own actions.
This is typically accomplished by first learning a generative model of likely solution steps
from human-written proofs \cite{whalen2016holophrasm, polu2022formal}.
However, there is little --- if any --- data from students writing
their solutions in existing theorem-proving languages\footnote{This would also be the case were we formalizing
and exploring a new mathematical domain, a setting where many researchers hope automated theorem provers could be useful.}. Therefore, we wish to proceed without this dependency.

To enable learning by selecting actions --- rather than generating them --- and
receiving a termination signal once a solution has been found,
we first need an environment where the set of valid next moves
is finite, while maintaining generality as much as possible.
To that end, we propose Peano, a simple theorem-proving language based on a dependent type system,
where the space of next proof steps is constructed to be finite.
We describe Peano in detail in Section~\ref{sec:peano}.
The Peano environment defines states as the sequence of terms
constructed so far. A solution is complete once it constructs a term
that satisfies the goal of each of the sections
(e.g., an equality between $x$ and a constant when solving an equation).
Peano allows us to easily implement all the axioms we need to
solve problems from the sections in Figure~\ref{fig:ka-sections}.
The environment can enumerate all the ways in which axioms can be applied to
the current state, thus giving us a set of actions at each state.
Finally, in section \ref{sec:case-study}, we implement simple problem generators for each of the domains by taking exercises
from Khan Academy and turning them into templates
(e.g., assuming the constants could be chosen at random),
which gives a distribution over initial states.
Together, these components (states, actions, a termination criterion, and a distribution
over initial states) yields a well-defined search problem.

While the set of actions at a given state is finite,
the combinatorial search space hinders unguided search algorithms
from solving all but the most trivial problems.
Thus, to make progress, we need to guide search using heuristics.
To maintain generality, instead of manually encoding domain-specific heuristics,
we would like to \emph{learn} these heuristics from past searches.
To that end, we start by training an agent using Contrastive Policy Learning (ConPoLe)
\cite{poesia2021contrastive},
a method to learn a policy for guiding search that has been shown to work on similar symbolic
reasoning domains with sparse binary rewards. 
To sample problems during training,
we first choose one of the sections at random, and then use that section's problem generator.
During learning, we evaluate the agent on a set of held-out problems from each section.

ConPoLe can learn to solve all problems in the first two sections, where solutions might
have up to 6 formal steps.
However, the third section already poses major challenges: solutions are both
longer (requiring up to 9 steps) and require combining a wider range of axioms.
The number of available actions also grows the more steps we add to the solution,
making solutions exponentially less likely to be found by exploration.
As a result, ConPoLe fails to make meaningful progress on the last three sections,
only succeeding at solving trivial equations ($x + 0 = k$, $1 * x = k$)\footnote{
Given a manually written solution, we can estimate the probability that a random
agent would generate that solution with a given action space by multiplying the
probabilities of picking the next action to match the solution.
For the shortest solution to $x + 1 = 2$, this probability is approximately $10^{-12}$.}.
This problem would grow even worse were we to continue progressing throughout the
Khan Academy curriculum further.
How would an agent possibly learn to solve problems from the third section and beyond?

A key insight becomes clear once we analyze how students are taught to solve equations
in these sections. When explaining how to solve $x + 1 = 10$, the instruction begins
by observing one can subtract a constant from both sides of an equality,
using that to get $(x + 1) - 1 = 10 - 1$. From here, the instructor no longer refers
to the base axioms
(e.g., commutativity, associativity) as operations one needs to apply. Instead,
they assume that the student can ``simplify'' each of the sides to get $x = 9$.
Many examples of how to ``simplify'' have been seen in the previous sections.
The key observation is that the solution would only have 3 conceptual steps, not 9,
\emph{if the agent had a high-level ``simplify'' action} (subtract from both sides,
simplify one side, then simplify the other).
With that action in the action space, the probability that the agent would find
solutions to learn from would be non-negligible.

How can the agent \emph{learn} such an action?
Since ``simplify'' is a generalization of what is learned from exercises in the
first sections, one alternative is to \emph{induce} that action by abstracting
steps from solutions found so far.
In Section~\ref{sec:tactic-induction}, we describe
a simple method based on anti-unification \cite{pfenning1991unification}
which can create high-level
actions by finding patterns in Peano solutions.
Following other theorem-proving languages, we
call these high-level actions \emph{tactics} ---
a procedure that can perform several operations, including calling other tactics,
to manipulate the proof state. Given the ability to induce tactics from its own
solutions, the agent is indeed able to make steady progress and eventually solve
all problems from the 5 sections.

Human mathematics poses another puzzle: after someone has mastered a mathematical domain,
the next generation of students can get to the same point much faster, to then go beyond.
Indeed, mathematical discoveries such as calculus and analysis took hundreds of years
to be developed and perfected; yet high-school and college students today 
often learn the core of these topics within semester-long courses.
What enables us to transmit mathematical knowledge so efficiently?
We propose that part of the explanation has to do with exploiting the structure
of abstractions that underlie the domain. More specifically, while it may take many
sparse experiences to realize that an abstraction is useful, mathematicians can then
collect those experiences so that the next generation can see it immediately,
almost as if the need for a particular concept was obvious in the first place.
The process of selecting and sorting examples for the next generation
is what we understand as \emph{developing a curriculum}.

We can induce a curriculum from our generated problems by analyzing
the agent's solutions and what tactics they use.
Since tactics can invoke either axioms or other tactics, we can arrange them in a dependency
graph. These dependencies imply a natural partial order on problems by considering
 which tactics are present in their solutions, combined with the dependencies between those tactics.
Sorting problems in topological order ---  respecting the dependencies between tactics
in their solutions --- then gives a curriculum.
Given an automatically constructed curriculum, two natural questions arise.
First, how similar is that curriculum to the human-designed ordering present
on Khan Academy? Second, can this curriculum serve an analogous purpose of
speeding-up learning of a ``second generation'' of agents?
We explore both questions in Section~\ref{sec:curriculum}.

\begin{figure}
    \centering
    \includegraphics[width=\textwidth]{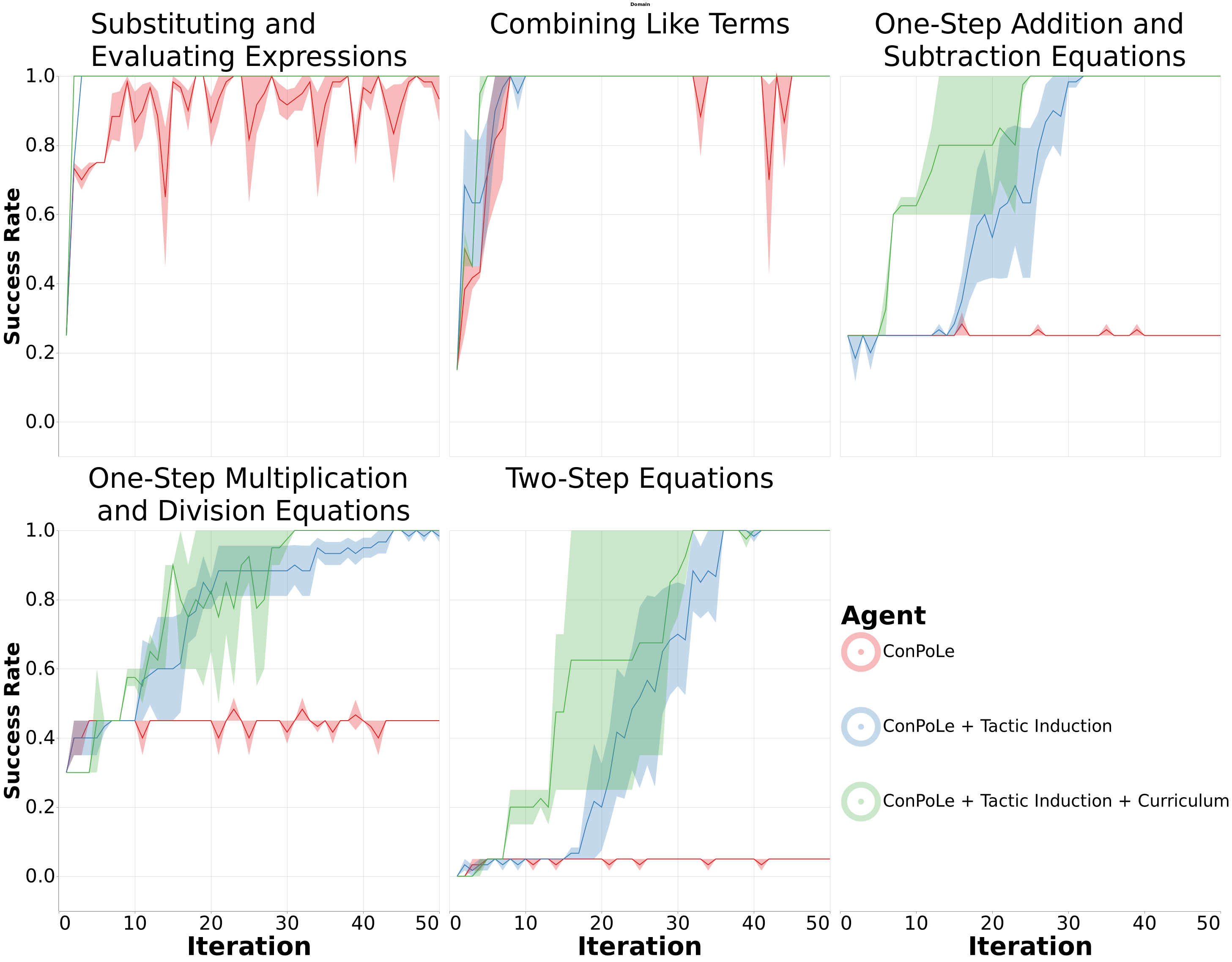}
    \caption{Learning curves of (a) a vanilla ConPoLe agent, (b) an agent with tactic induction,
    and (c) a ConPoLe agent with tactic induction trained on the curriculum induced from the previous agent.}
    \label{fig:learning-curves}
\end{figure}

\section{Peano: Theorem Proving with a Finite Action Space}
\label{sec:peano}

In this section, we aim to define a theorem proving environment that presents an agent with a finite action space,
where each such action corresponds to adding a solution step.
Our goal is to obtain (1) a simple yet expressive language to specify
mathematical domains (their definitions, axioms, problems and solutions), and simultaneously
(2) an environment where an agent can solve problems formalized in the language by
sequentially choosing among a finite set of valid solution moves.

Many theorem-proving languages, including Coq, Isabelle/HOL and Lean, would already fulfill
our first requirement. These languages are based on a relatively small core
(dependent type systems in Coq or Lean, or higher-order logic in Isabelle/HOL) and they have
been used to fully formalize extremely high-level mathematical results (e.g., the independence of
the Continuum Hypothesis and the proof of the Kepler Conjecture).
But when seen as environments for agents to find proofs, these languages pose challenges in how
a search space with a finite branching factor might be defined.
This is the main challenge we address here.

To overcome the unbounded space of valid solution steps, several prior works have used generative 
models trained on datasets of human-written formal proofs as a way to sample sensible actions in
a given context. This approach works in the presence of many existing formal proofs for training,
but here we would like to proceed without this assumption.
Thus, to circumvent the need for the agent to \emph{generate} valid actions,
we want to design an action space so that solution steps can be simply \emph{selected} out of a finite
set of options.

We start by noticing that simply making a finite and complete action space would be a vacuous goal
if actions do not correspond to complete, valid solution steps.
For instance, we could trivially consider character by character generation of proofs in any
existing theorem-proving language, with a special action to mark the end of the solution.
Once that character is chosen, we can check whether the solution was valid by submitting it to the
language's verifier.
While finite, an untrained agent has virtually no chance of generating valid solutions in this space,
making learning from sparse rewards implausible.
Leveraging the language's grammar improves this situation by eliminating lexical and syntactical
errors. However, since most of the constraints are imposed by the language's semantics,
sampling from the grammar will still produce ill-formed solutions with overwhelmingly high probability.
Instead, we would like to completely rule out semantically invalid solutions,
and define a space where all steps are always complete and valid, though
they might not necessarily move towards the goal.

\subsection{Foundation: $\lambda$-calculus and dependent types}

To define solution steps, we first need a logical foundation to represent definitions,
propositions, axioms and solutions.
We find the language of \emph{dependent $\lambda$-calculus} \cite{martin1984intuitionistic} to be a compelling choice for our purpose.
Dependent $\lambda$-calculus borrows the three basic constructors of terms from $\lambda$-calculus:
a term can be a variable (e.g., $x$), a function application (e.g., $(f\ t_1\ \cdots\ t_n)$)
or a $\lambda$-abstraction (e.g., $\lambda x : T. t$).
Each term is always associated to a \emph{type}, which essentially constrain which terms can be
used as arguments in function applications.
Most type systems have separate formation rules for types and for objects of those types.
Dependent type systems blur this distinction, allowing types themselves to be terms that
depend on the \emph{values} of other terms.
As we explain below, this device makes dependent types
a rather succinct language to express many common patterns of mathematical
reasoning in a small foundation.
Moreover, having a type system in our environment will let us drastically narrow down the set of
actions available to an agent by leveraging type constraints.
This contrasts to untyped foundations such as first-order logic
or ZFC Set Theory, where the constraints imposed by the formal system itself are mostly syntactic,
and user-defined predicates encode most of the semantic constraints
\footnote{For instance,
  classical first-order logic assumes a single underlying universe of objects.
  In an interpretation where objects include both numbers and sets,
  if we introduce a function symbol $U(s_1, s_2)$ for the union between two sets,
  it is in principle valid to consider the union of two numbers.}.

In a language to represent mathematical reasoning,
we wish to represent both typical mathematical objects (e.g., the number 2)
and propositions about those objects (e.g., that 2 is even).
Dependent types can encode both notions (of \emph{kinds} of objects and
\emph{propositions} about those objects) in a unified manner.
Two base types are provided in Peano, called \texttt{type}
and \texttt{prop}.
Regular types, such as the natural numbers, are themselves terms of type \texttt{type},
while propositions are encoded as terms of type \texttt{prop}\footnote{
In our current implementation of Peano, the type of both \texttt{type}
and \texttt{prop} is \texttt{type} itself.
This does lead to Girard's paradox, but this is not a practical issue for our experimental investigation.
Lean and Coq circumvent this issue by defining an infinite sequence of types \texttt{type i}
where each \texttt{type i} is of type \texttt{type (i + 1)}, a solution we can also apply.}.
In a dependent type system, terms can themselves be (part of) types of other terms.
For example, suppose we define a type \texttt{nat} to represents natural numbers.
The proposition that a natural number $n$ might be even can be encoded as a function
with input type \texttt{nat} (the number $n$) and output type \texttt{prop}
(the proposition that $n$ is even)\footnote{Note that dependent types allow for more than simple parametric polymorphism,
where types can depend on other types but not on regular objects.
A typical example of the latter would be the polymorphic type \texttt{List<T>}
where each other type \texttt{T} yields a different list type.}.
Constructors for proposition types, such as the types produced by \texttt{even}, define how proofs of those
propositions can be created.
Note that there is a distinction of creating the \emph{proposition} ``$n$ is even''
and an object of type ``$n$ is even'' --- the former does not imply that the latter is possible.
Indeed, we would like the system to never allow a construction of an object of type ``$3$ is even''.
In this language, proofs are represented as programs: to prove a proposition, we
must give a program that constructs an object of corresponding type when given objects
of the hypothesis types.
For example, when given a natural number $n$ and an object of type ``$n$ is even'', we might
be able to construct an object of type ``$n^2$ is even''.
This encoding of propositions as types and proofs as programs is the well-known
Curry-Howard correspondence.

Typically, a type system is designed to enable the implementation of a type-checker for a formal language:
given a program $P$, the type-checker verifies whether $P$ satisfies the type constraints.
When $P$ encodes a proof, this corresponds to checking the validity of the proof.
Our main use of a type system is different: we'll use the rules of the system to define an
action space to \emph{generate} valid programs.
This will also let us easily check the validity of  proofs, however, by sequentially checking whether each step could have been generated from the action space.

\subsection{Defining a background library}

When using interactive theorem proving environments, we usually have a distinction between
a global \emph{background library}, which has axioms and other theorems that we might use
and is fixed while proving a particular theorem, and a local, dynamic \emph{state}
which the solution steps can operate on.
Both the library and the state can be represented as sets of typed objects.
In Peano, we can write:

\begin{verbatim}
nat : type.
z : nat.
succ : [nat -> nat].
one : nat = (succ z).
\end{verbatim}

This syntax is inspired by Elf \cite{pfenning1994elf}.
These statements declare a type called \texttt{nat}, then an object of type \texttt{nat} called \texttt{z},
a function called \texttt{succ} which receives one \texttt{nat} and gives another \texttt{nat},
and an object called \texttt{one}, again of type \texttt{nat}, which is defined as an alias for the application
of \texttt{succ} to \texttt{z}.
At this point, our library would have 4 objects.
Note that \texttt{succ} is an uninterpreted function:
all we know is its type signature, but we do not need to provide an implementation.

These same constructs can be used to define properties about \texttt{nat}s, such as the
``less than or equal to'' (\texttt{leq}) relation:

\begin{verbatim}
leq : [nat -> nat -> prop].
z_leq_n : [('n : nat) -> (leq z 'n)].
n_leq_sn : [('n : nat) -> (leq 'n (s 'n))].
\end{verbatim}

Thus, \texttt{(leq one z)} is a proposition
(which doesn't imply it is true, since we have not constructed any object of that type).
\texttt{z\_leq\_n} is an axiom: if we apply it to any natural number \texttt{'n}, it outputs
an object of type \texttt{(leq z 'n)}. Here, we named the parameter (as \texttt{'n})
since the output type depends on it. A single quote is used to prefix bound variables,
in order to easily distinguish them from variables coming from the background library.
Parameter types can also introduce variables that are defined at the time of
function application by unification:

\begin{verbatim}
leq_trans : [(leq 'a 'b) -> (leq 'b 'c) -> (leq 'a 'c)].
\end{verbatim}

Thus, if we have proofs that $a \leq b$ and that $b \leq c$
for some $a, b$ and $c$, this axiom of transitivity would let us obtain a proof
that $a \leq c$. We could also have written this axiom in a longer form,
by taking three \texttt{nat} parameters separately and then the two
mentioned proof objects.
This choice is not merely syntactical: while the two
versions will allow the same actions, the version where
$a$, $b$ and $c$ must be inferred will
enable much more efficient action enumeration
(Algorithm~X, described below).

In addition to user-defined types and axioms,
Peano provides a built-in dependent equality type
and three axioms to handle equality: \texttt{eq\_refl} encoding reflexivity
(from any type $T$ and object $t$ of type $T$,
we can construct evidence that $t = t$), \texttt{eq\_symm} encoding symmetry (from two objects $a$ and $b$ and
evidence that $a = b$, we can construct evidence that $b = a$)
and \texttt{rewrite} encoding congruence (for any object of a proposition type $P$ and
an equality $a = b$, we can construct an object of type $P[a \mapsto b]$, i.e.
the result of substituting one free occurrence of $a$ by $b$ in $P$).

\subsection{Problems, states and goals}

Having defined a background library, we would like to pose problems and construct solutions to them.
In Peano, the solution state consists of a set of objects that have been constructed so far
--- including proofs of intermediate propositions.
A problem is defined as a set of starting objects --- 
for example, two \texttt{nat}s $a$ and $b$ and a hypothesis \texttt{(leq a b)} ---
and a goal. Our main deviation here from other proof assistants is in how goals are specified.

In most interactive proof assistants, the prover typically starts by
writing down a proposition to be proven --- a type, which is taken as an open goal.
Then, a series of proof steps will follow, each might change the proof state or the open goals;
the prover is done once there are no more open goals.
Essentially, these systems close an open goal once they construct an object of the goal type.
Once no more open goals remain, the proof assistant has enough information to
construct an object of the original proposition type.

Several common classes of educational exercise do not fit into this paradigm in
a straightforward manner.
In particular, many exercises, such as CLT in Figure~\ref{fig:ka-sections},
involve manipulating an expression until a syntactic condition is met
 --- for instance, until we obtain an equivalent expression that is simplified,
 or an equation with a variable on the left-hand side and a constant on the right side.
These problems require checking assertions about syntactic forms, rather than the
objects they represent (e.g., an expression is simplified, or an equation is solved).
To reason about properties of expressions in the type system,
we would need to reify the \emph{expressions} themselves, and model the relation
between expressions and the objects they \emph{denote}.
This is certainly possible within a dependent type system, and brings conceptual clarity
to what exactly are the problems at hand,
but requires adding another layer to the encoding of problems and solutions
in the formal system.

To circumvent this complexity in the context of simple educational domains,
we allow a more flexible approach of simply having a small program,
written outside of the formal system, that checks if the given solution meets the goal.
This program receives the state --- all constructed objects and their types ---
and returns a binary reward signal.
Under this flexible model, the usual goal of constructing an object of a certain type
--- a proof of a proposition --- is trivial to check with a program that searches the state
for an object that has the goal type.
But syntactic conditions also become easier to test. For example, to check if
``an equation is solved'', a verifier can
check if the type of any of the constructed objects is an equality between $x$ and a constant.
If the prover managed to produce such an equality, then we can say it managed to ``solve the equation'',
since the formal system will guarantee that all steps that have been taken leading
to this equality were valid.

Thus, in Peano, the state consists of a set of typed objects, a problem specifies
the starting state and the goal of a problem is to modify the state until it
satisfies a known verifier. In other words, the prover has the information of whether
the goal is to ``solve the equation'' or to ``prove a particular proposition''.
We now look at which actions the prover can take to change the solution state.

\subsection{Finite action space}

In modern dependently-typed proof assistants, such as Lean or Coq,
the proof writing process typically enables an infinite set of valid proof steps at any given point.
We first identify the sources of infiniteness, and then describe how we constrain this space to be finite.

When processing a proof, existing proof assistants typically keep a state consisting of a set $S$ of objects
that have been constructed or are assumed to exist (starting with the hypotheses)
and a set $G$ of goals. The goals $g_{i} \in G$ are propositions --- thus,
types --- and we can complete our proof once $S$ contains objects of each of the types in $G$.
A proof step can therefore have effects in either $S$ or $G$.
Changes to $S$ are \emph{forward reasoning steps}: they typically construct a new
object, which is added to $S$.
The two essential means of constructing objects come from $\lambda$-calculus:
function application and lambda abstraction.

Proof steps might also change $G$: these are \emph{backward reasoning steps}.
These proof steps also essentially consist of applying
an existing function $f_{b}$, but this time \emph{in reverse}
to a goal $g_{i}$. In this case, $f_b$ must output the goal type,
and its argument types then replace $g_{i}$ in $G$ as new sub-goals to be closed.
If we eventually satisfy the sub-goals by constructing objects of the appropriate types,
an object of type $g_{i}$ --- to satisfy the original goal --- could be produced by applying $f_{b}$ in the forward direction.

Both the sets of forward and backward steps might be infinite.
The number of valid forward steps ---
either function application or lambda abstraction --- is unbounded because
a single step can construct arbitrarily deep objects.
For example, if we have the natural number $0$ and the successor function
$S : \mathbb{N} \rightarrow \mathbb{N}$,
a single forward step might construct any natural number\footnote{In Lean,
  one could write \texttt{let n : nat := (s (s (... z))).}}.
  Similarly, a lambda abstraction can have an arbitrarily complex body.
  We can easily constrain function application steps without sacrificing
  generality by forcing
  each step to apply a single function to a combination of already existing arguments.
  Deep objects can still be incrementally constructed in multiple steps.
  A similar --- but more subtle --- idea could be applied to allow the construction of
  lambda abstractions with a finite set of actions at each step.
  Our current version of Peano does not include the additional actions needed to inline lambda abstraction.
  This limits our current action space to only produce proofs
  that don't require auxiliary functions (or lemmas).
  When using Peano, one can still prove theorems that require lemmas by stating and proving the
  lemmas separately (i.e., not inline in the main proof).
  This makes some proofs unnatural to write\footnote{For example, in a proof by induction,
    each of the branches is a lemma that currently has to be outlined.},
  but does not impact the educational mathematical domains we study in the present work.

  Our relaxation of goals into general verifiers (that determine if a state is ``done'' beyond the existence of objects of certain types) complicates the specification of backward steps.
  However, though backward steps add flexibility to how one can construct a proof, they do not fundamentally enable new proofs --- any proof that contains backward steps can be mechanically
  translated into a proof that only uses forward steps.
  Furthermore, human students typically start using backward steps at a later level
  of education than we consider in our algebra domains. Thus, we also limit ourselves to forward steps. Together with the above constraint on forward steps,
  this makes the action space finite.
  We note that a strategy similar to how we constrained forward steps
  could also be used to arrive at a finite set of backward steps while still retaining
  generality\footnote{Specifically, we could
    allow a backward step when all the produced sub-goals can be expressed with
    already existing objects. Multiple steps might then be needed to construct the necessary objects to
    apply a backward step.
    This would overcome the fact that, in proof assistants,
    backward steps typically allow the immediate, arbitrary instantiation of existential quantifiers
    as long as the result type-checks, thus allowing infinite options.}.

  In summary, the action space in Peano contains all valid forward steps that apply a function
  to a combination of existing objects, thereby creating a new object.
  The type system determines which objects are allowed as arguments,
  and this constraint lets us efficiently enumerate all of the finitely many available actions.
  Algorithm~\ref{algo:actiongen} describes this procedure more concretely.
  The algorithm is essentially a backtracking search for all ways to fill in arguments of a chosen
  function (e.g., an axiom) with objects coming from a given collection $S$.
  Because of dependent types, choosing a value for an argument
  might change the expected type of later arguments (e.g., a function could first take a number $n$
  and then a proof that $n$ is even; filling in a concrete value for $n$ thus determines a concrete type
  for the other argument).
  To perform one step in a solution, the agent first chooses a function to apply.
  Then, we invoke Algorithm~\ref{algo:actiongen} to enumerate the results that can be obtained
  with the chosen function. Finally, if the result set was not empty, the agent chooses one of those results to add to its current solution.

  \begin{algorithm}
    \caption{Action enumeration in Peano}\label{algo:actiongen}
    \begin{algorithmic}[1]
    \Procedure{EnumerateActions}{$S,\ f$}\Comment{Computes all results of applying the given}
    \State \textbf{return} \textsc{FillNextArgument}$(S, f, \langle \rangle, f.param\_types())$\Comment{function $f$ with parameters}
    \EndProcedure \Comment{coming from the set $S$.}

    \Procedure{FillNextArgument}{$S,\ f,\ a, param\_types$}
        \If{$|a| = f$.number\_of\_arguments()}
            \State \textbf{return} $\left\{ f(a_1, \cdots, a_n) \right\}$
        \EndIf

        \State $next\_type \gets param\_types[|a|]$
        \State $results \gets \varnothing$
        \ForAll{$obj \in S$}
            \State $mgu\gets unify(obj.type, next\_type)$
            \If{$mgu \neq null$} \Comment{Check if $obj$ can be used as the next argument.}
                \State $a' \gets \langle a, obj \rangle$
                \State $new\_param\_types \gets param\_types$
                \ForAll{$(var, assignment) \in mgu$} \Comment{Make substitutions in later parameter types.}
                    \State $new\_param\_types \gets \sigma(new\_param\_types, var \mapsto assignment)$
                \EndFor
                \State $results \gets results\ \cup\ $ \textsc{FillNextArgument}$(S, f, a', new\_param\_types)$
            \EndIf
        \EndFor
        \State \textbf{return} $results$
    \EndProcedure
    \end{algorithmic}
  \end{algorithm}

\section{Case study: solving Khan Academy problems in Peano}
\label{sec:case-study}

The main design goal of Peano is to provide a flexible representation
for educational mathematical domains, both to let us understand how learning
in this domains can take place and to facilitate applications in computer-assisted education.
We now describe at our main case study in this direction: the formalization
of the algebra sections of Khan Academy illustrated in Figure~\ref{fig:ka-sections}.
These sections assume a student who starts with the knowledge of how to
evaluate the basic operations with known real numbers, as well as various
properties of these operations (e.g., commutativity and associativity).
From there, they teach this student to solve simple linear equations with one unknown.

To describe our formulation of these sections in Peano, we must specify
the basic definitions --- types and axioms --- how we generate problems
and finally how we specify goals.

\begin{figure}
    \centering
    \includegraphics[width=\textwidth]{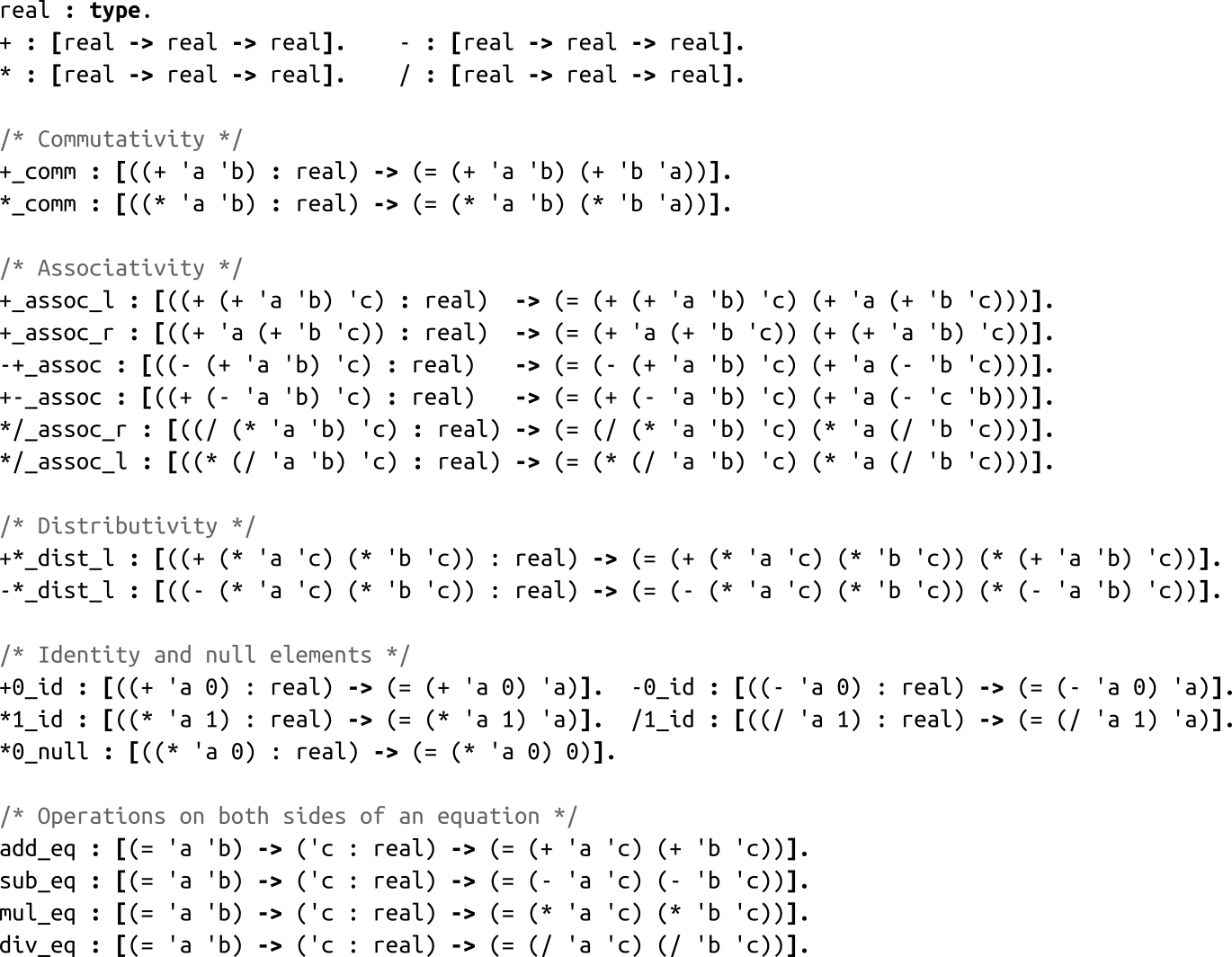}
    \caption{Axioms in Peano used to formalize the sections of Khan Academy that we study.}
    \label{fig:ka-axioms}
\end{figure}

\paragraph{Definitions and axioms}
Figure~\ref{fig:ka-axioms} shows the full Peano representation of the axioms we use.
All axioms here output equalities relating terms in their inputs.
Most parameters to the axioms are real numbers,
but their values in most cases need to be the result of an expression of some form.
For instance, the parameter to \texttt{+\_comm}, the commutativity of addition,
is a number of the form \texttt{(+ a b)} for some $a$ and $b$.
When applied to an argument of the necessary form, the output type of
\texttt{+\_comm} will be the equality type \texttt{(= (+ a b) (+ b a))}.
Note that we can equally formalize this (and all other) axioms without
pattern matching on syntactic forms, but rather by taking \emph{any}
numbers $a$ and $b$ and returning an equality \texttt{(= (+ a b) (+ b a))}.
While this latter form would more faithfully represent the property
of commutativity of addition, it would also generate more actions during proof
search, since it applies to any real numbers regardless of whether we're
already considering their sum. Thus, the form we write these axioms 
reflects both the properties we need and
the fact that we tend to apply these properties once we already have
an object that suggests they will be useful.
In practice, this narrows down the action space without ruling out natural solutions.

\paragraph{Evaluating expressions}
The four binary operations on reals are declared as uninterpreted
functions. To focus on algebraic reasoning, we add an additional axiom \texttt{eval},
implemented outside of the formal system, which can be used to ``execute'' these
operations when their arguments are known. From the agent's perspective,
\texttt{eval} is an axiom which takes an object of type \texttt{real}
as its only parameter and,
if that real is of the form \texttt{(op a b)}, where \texttt{op} $\in \{+, -, *, /\}$
and both $a$ and $b$ are constants, \texttt{eval} returns an equality proof between
\texttt{(op a b)} and the result of evaluating the expression (for example:
\texttt{(eval (+ 1 2))} would have output type \texttt{(= (+ 1 2) 3)}).
If we were to fully formalize the content on Khan Academy, it would be more faithful
to not have an atomic evaluation procedure, but rather break it down into more basic
steps, corresponding to how students are taught arithmetic.
For our current case study, however, we assume evaluation is given as atomic.

\paragraph{Generating problems}
For each of the sections in Figure~\ref{fig:ka-sections}, we create random
problem generators by creating syntactic templates with placeholders that are
then replaced with random numbers.
We list these templates in Table~\ref{tab:problem-generators}.
Some of these templates come directly from exercises from Khan Academy ---
for example, the ``one-step equation'' $x + 10 = 27$ turns into the template
\texttt{(= (+ x n1) n2)}, where $n1$ and $n2$ can be replaced by any constant. 
Since the pool of exercises on Khan Academy is fixed and small
(4 to 7 exercises in each practice section), 
we add a few templates to increase the diversity of problems generated
and ensure that they employ all axioms.
We generate integer constants by rounding samples from a Gaussian $\mathcal{N}(0, 25)$,
and ensure that we do not sample divisions by zero or absurd equations such as $0x = 1$.
For problems that involve solving an equation, we declare a real number named \texttt{x}
and an object named \texttt{equation} whose type is the equality type corresponding
to the equation to be solved. For the first two sections, which involve simplifying
an expression in some way, we declare a real number named \texttt{answer} and
encode the problem by assuming an equality between \texttt{answer} and the
expression to be simplified.

\begin{table}
    \centering
    \begin{tabular}{c| c}
        \textbf{Section} & \textbf{Syntactic forms used in problem generator} \\
        \hline
        SEE & \texttt{(= x (+ 1 2))}, \texttt{(= x (* (+ 1 2) 3))}, \\
            & \texttt{(= x (+ 1 (* 2 3)))}, 
              \texttt{(= x (/ (* 1 2) (- 3 4)))} \\ \\
        CLT & \texttt{(= answer (+ (- x 1) 2))},
              \texttt{(= answer (- (+ x 1) 2))} \\
            & \texttt{(= answer (* (/ x 1) 2))},
              \texttt{(= answer (/ (* x 1) 2))} \\ \\
        OAE & \texttt{(= (+ x 1) 2)}, \texttt{(= (- x 1) 2)}  \\ \\
        OME & \texttt{(= (* x 1) 2)}, \texttt{(= (* 2 x) 3)}  \\
            & \texttt{(= (/ x 2) 4)} \\ \\
        TSE & \texttt{(= (+ (* x 2) 1) 3)}, \texttt{(= (- (* x 2) 1) 3)}, \\
            & \texttt{(= (+ (/ x 2) 1) 3)}, \texttt{(= (- (/ x 2) 1) 3)} \\
    \end{tabular}
    \caption{Syntactic forms used in our problem generators for each of the
    Khan Academy sections. Each problem is generated by picking one syntactic
    form uniformly at random and then randomizing constants.
    Furthermore, in Substitution and Evaluating Expressions (SEE), all
    operations are resampled from $\{+, -, *, / \}$.}
    \label{tab:problem-generators}
\end{table}

\paragraph{Goals}
Having problems and axioms, the last step in formalizing the domains we study
is to define what it means for a solution to be complete.
For the first two sections, where the goal is to simplify the expression,
our solution checker searches the state for an equality between
\texttt{answer} and an expression in a simplified form (we enumerate a few
forms that cover the exercises we formalize: \texttt{c}
(a constant), \texttt{x}, \texttt{x + c} with $c \neq 0$, and variants
where $x$ is multiplied by a constant that is neither $0$ nor $1$).
When the goal is to solve for $x$, our checker searches for
an equality of the form $x = c$ for some constant $c$.
Once the solution state satisfies the check corresponding to the problem,
the solution is taken as complete.

\begin{figure}
    \centering
    \includegraphics[width=\textwidth]{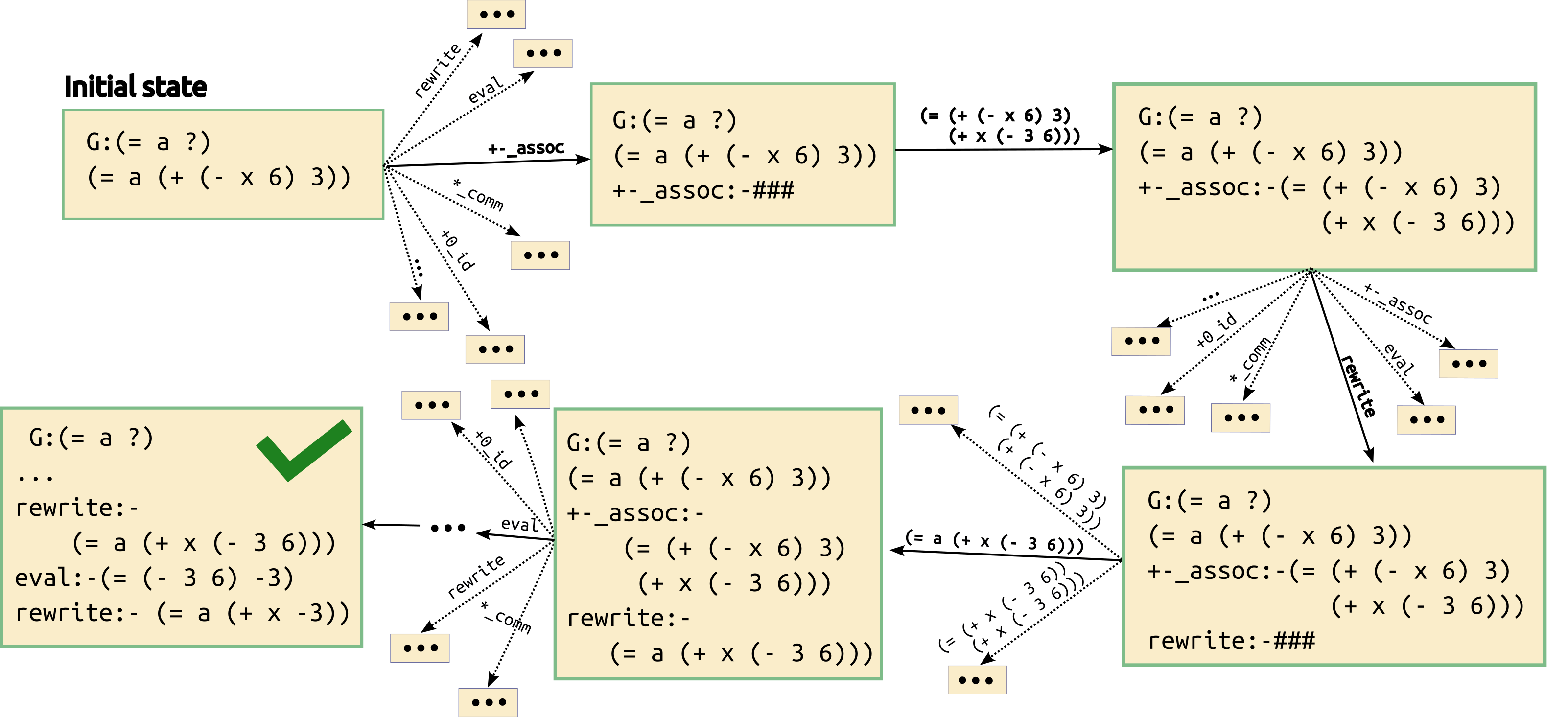}
    \caption{Example of state and action sequences solving a problem from the Combining Like Terms section. Here, the goal, represented in the state by a short line, is to find a simplified form for the variable \texttt{a}. At each state,
    the agent needs to either select one axiom from the library or choose one the results from that axiom to be added to the state. The solution proceeds by applying the appropriate associativity rule, then using that to rewrite a, then evaluating the resulting operation with constants, and doing a final rewrite to arrive at a solution state.}
    \label{fig:search-space}
\end{figure}

\paragraph{Example}
Figure~\ref{fig:search-space} shows an example of solving a problem generated
in the Combining Like Terms section. The solution uses 4 axiom applications:
first the applicable case of associativity, then rewrite (the equality axiom corresponding to congruence), followed by evaluation and a final rewrite.
At that point the goal is satisfied, since the state contains an equality between
the ``answer'' variable and a fully simplified expression.
The probability that this solution would be generated by an agent picking
a sequence of actions at random is $9.64 \times 10^{-7}$,
which we compute by multiplying the inverse
of the number of available choices at each state.
Thus, even simple problems have a non-trivial search space when
a modestly-sized library is available.

\section{Learning to solve mathematical reasoning problems}
\label{sec:learning}

Given the ability to sample problems, enumerate and apply actions, and to
detect solution states, the problems from Khan Academy yield a
family of well-specified search problems.

Any search algorithm can be applied to attempt to find a solution state given a problem.
But general, domain-agnostic search methods
are constrained to finding extremely short solutions, as illustrated
by the example in Figure~\ref{fig:search-space}.
This explosion grows worse the deeper we explore, since each action adds objects
to the state, which in turn enable a growing number of new objects to be constructed
by applying axioms.
Thus, to be able to find solutions to harder problems, we need to leverage
heuristics to guide search.

How can we encode problem-solving heuristics?
One option is to manually design some of these search strategies,
such as preferences for certain actions given some features of the state.
But this approach involves
significant engineering that must be repeated for every new domain we wish to formalize.
Moreover, we still risk missing cases and developing incomplete
strategies: even in simple-looking domains such as equation solving,
expert-written heuristics for finding solutions in this step-by-step fashion
risk reaching failure cases when tested more widely, as demonstrated in
\cite{poesia2021contrastive}.

Instead of relying on domain-specific design, we aim to \emph{learn} search heuristics,
bootstrapping from easy to hard problems by learning from past searches.
One form of encoding search heuristics is through a \emph{policy}: a function
that takes states and gives a probability distribution over actions.
A search algorithm can then leverage a policy as a heuristic to prune unlikely actions during search.
To learn a good policy, Expert Iteration (ExIt; \cite{anthony2017thinking}) provides a simple general
paradigm: we can alternate between running search on batches of problems using our current policy and
training the policy by imitating decisions made during previous successful searches.
When applied to deterministic problems with a binary reward signal, it is typical to
ignore unsuccessful attempts and train only on data from successful searches \cite{polu2022formal,poesia2021contrastive}.

One instance of this paradigm that we can directly apply to our
current setup is Contrastive Policy Learning (ConPoLe; \cite{poesia2021contrastive}).
ConPoLe was introduced as a method for policy learning to solve
symbolic reasoning problems from the Common Core environments,
which include equation solving and fraction simplification.
The Common Core environments include domain-specific representations
for states and actions which make them a less general setting than Peano.
For example, they embed the assumption that, when solving a single equation,
one can forget about all steps except for the last.
This assumption severely reduces the state space, but does not generalize
(e.g., even to systems of equations, where we often change which equation we're working on).
Nevertheless, these environments also present the challenge of learning from
unstructured text representations and sparse binary rewards,
making ConPoLe a natural choice to try on the Khan Academy problems.

The main idea of ConPoLe is to apply a search algorithm that can use
a policy $\pi(a|s)$ to prioritize search nodes --- for instance, beam search ---
to a sampled batch of problems.
Then, the solutions found (typically few and short at the beginning)
are used to train $\pi$ by a reduction to contrastive learning:
ConPoLe learns a representation $\phi(s)$ for states that attempts to align each
state with a successor leading to a solution, using all other successor states
available during search as negative examples.
Thus, ConPoLe is compatible with our setting, where the state and action spaces
are unbounded, the set of available actions depends on the state,
the effects of actions are deterministic and only a sparse, binary reward signal
is given once a solution state is reached.

To apply ConPoLe to the Peano formalization of Khan Academy problems,
we simply need to represent states and actions as strings, and define a
differentiable neural architecture for the embedding function $\phi$.
To represent $\phi$, we use a character-level, two-layer bidirectional GRU network \cite{cho2014properties}.
We encode states by first formatting the initial objects given in
the problem (e.g., the given equation), then formatting the objects constructed
by each action taken so far. We truncate the state at the beginning to
bound the number of characters the state embedding function might receive
at 200 characters.
For actions, we found that enumerating and running all possible applications of
axioms through $\phi$ to be slow, as deeper states can have hundreds
of such actions available. To mitigate this problem, we decomposed the solution
generation process into pairs of actions: first, the agent chooses which
axiom to apply; then, the Peano runtime enumerates the valid applications
of that axiom in the current state, and the agent chooses between one of the
achievable results to add to the current state.

\subsection{Result: learning to solve by pure policy learning}

Figure~\ref{fig:learning-curves} (red curve, ``Bare agent'') shows
the results we obtained when applying ConPoLe to the Khan Academy problems in Peano.
Here, the agent is being trained by sampling and attempting problems in a random order,
following the setup from \cite{poesia2021contrastive};
after every batch of $500$ problems, we train the policy using ConPoLe on examples generated
from successful training episodes, and evaluate
it on held-out problems from each of the Khan Academy sections.

ConPoLe makes steady progress in learning the first two sections
(Substitution and Evaluating Expressions and Combining Like Terms), eventually
learning to solve all problems in them. However, it stagnates in the equation solving sections.
We note two difficulties that are unique to our current setup when compared to
the Mathematics Common Core environments, in which ConPoLe succeeds in finding solutions to similar
equations \cite{poesia2021contrastive}.

First, the action space in Peano is significantly larger; a problem that compounds when
solutions get longer and thus more constructions are possible given all the constructed objects.
To give a sense of scale, the likelihood of a random policy solving a ``one-step addition equation''
from Khan Academy in the Common Core equations environment is $2 \times 10^{-5}$.
In contrast, in our Peano formalization, this likelihood drops to $10^{-12}$.
This difference alone severely affects the probability that any given policy will succeed
when solving new problems, since those require exploration in a large, combinatorial space.

Lastly, we observe that \emph{representation learning}, to which ConPoLe reduces policy learning,
has an additional challenge in Peano. Since states accumulate the results of previous steps,
it's much rarer to arrive at the same state twice.
In contrast, in the Common Core environments, since an action operates directly on the equation
at hand, actions might directly bring us to equations that we have seen before.
For example, once we simplify the right-hand side in $x + 1 = 1 + 2$ and arrive at $x + 1 = 3$,
it might be possible that our existing policy recognizes this new equation and can lead to a solution.
In Peano, this recognition that we can follow a previously discovered strategy is
more difficult, since the new state will also include the equality $1 + 2 = 3$ as well as
the previous equation. Thus, the representation of the new state might not immediately
tell us that we're essentially at a problem we have seen before.
In short, some domain-specific state abstraction is present in the Common Core environments
that instead has to be learned in Peano, making policy learning significantly more challenging.

\section{Tactic Induction}
\label{sec:tactic-induction}

Given any starting set of axioms, the number of deductions that can be made from them
typically grows exponentially as we enumerate longer sequences of steps.
Using higher-level axioms to attempt to make solutions shorter only delays this
challenge, but does not eliminate it.
Long chains of reasoning are also unwieldy for humans,
but the deductions that interest us
can be typically factored into higher-level steps,
so that even solutions to hard problems are made succinct in terms of
lemmas or procedures at an adequate level of abstraction.
To allow users to encode these abstract actions,
interactive theorem-proving languages often provide a language for \emph{tactics}:
programs that manipulate the proof state or goals that can encode higher-level
proof steps for a given domain.
A tactic can take parameters and invoke axioms, theorems or other tactics, ultimately
generating a potentially long sequence of actions in the underlying formal system.

To operationalize the idea that agents should learn high-level actions from experience,
we propose a simple tactic language for Peano.
A Peano tactic $t$ is a sequence of $n = |t|$ actions $t_a^{(i)}$,
along with action arguments $t_{p}^{(i)}$.
Each action $t_a^{(i)}$ is either an axiom or another tactic
(tactics do not yet support recursion).
Each $t_{p}^{(i)}$ is a list of arguments passed to action $t_{a}^{(i)}$.
In turn, each one of $t_{p}^{(i1)}, \cdots, t_{p}^{(ik)}$
might be either a concrete value or a symbol that references one of the
formal parameters of $t$ itself.

A tactic $t$ can be executed given a set of objects $S$, producing
a set of \emph{traces}.
Each trace corresponds to one valid combination of arguments for $t$.
To compute the set of traces that $t$ can generate given $S$,
we execute each of the actions in $t$ in sequence and lazily decide
which arguments for $t$ are possible given the objects that each action generates.
More precisely, for each action $a_i$, we first compute the set of valid arguments and results
that it can generate --- this will either call Algorithm~\ref{algo:actiongen}
for invoking axioms or call the tactic execution procedure recursively when 
$t_a^{(i)}$ is itself another tactic.
Then, for each choice of list of arguments that we can invoke $a_i$ with,
we unify that list with the existing assignments for parameters of $t$
in the current trace. If we find inconsistencies, we give up on the current
trace. If there are multiple possible assignments, then we \emph{branch}
the trace and continue execution in each branch. After all actions,
each of the resulting traces will correspond to one way of invoking
$t$, potentially producing multiple possible results. 

\begin{figure}
    \centering
    \includegraphics[width=\textwidth]{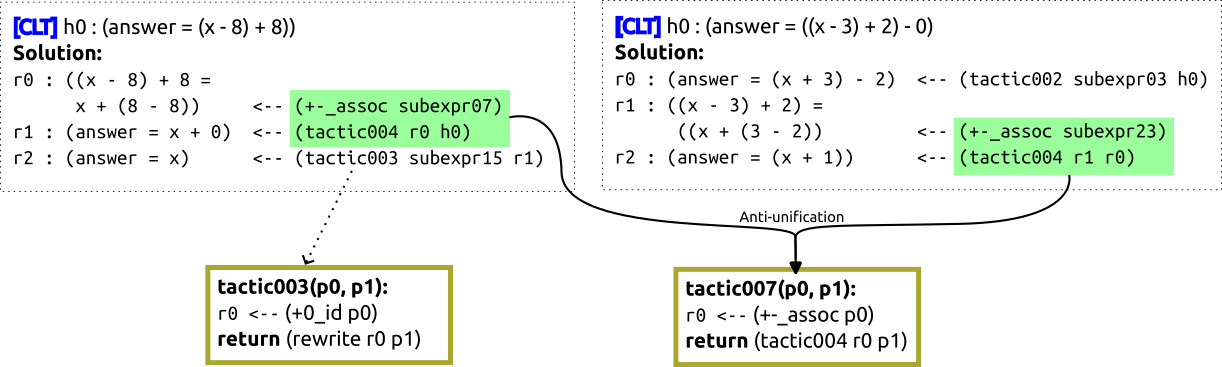}
    \caption{Example of two Peano tactics: \texttt{tactic003} simplifies an expression by invoking the axiom that asserts that $0$ is the identity element
    of addition, then using that equality to rewrite the original expression; \texttt{tactic007} is being induced by generalizing two segments of
    solutions to problems in Combining Like Terms.}
    \label{fig:tactic-example}
\end{figure}

Figure~\ref{fig:tactic-example} shows an example of two solutions in Peano
invoking a tactic named \texttt{tactic004}.
Here, this tactic represents the simplifying action
of applying the axiom that $0$ is the identity of addition,
and then using the resulting equality,
$\texttt{p0} + 0 = \texttt{p0}$ for some parameter $\texttt{p0}$, to rewrite $\texttt{p0} + 0$ to $\texttt{p0}$ in some other object $\texttt{p1}$.
Just like axiomatic actions, a tactic might generate one, multiple, or no results given the current
state: when the state consists of the inequality $e^{x + 0} < y + 0$,
\texttt{tactic004} generates two results;
and none in $x + 9 = 4$. When present in the action space, a tactic can be used by the agent like
any other action, as we described in Section~\ref{sec:peano}: the agent might first decide to invoke
the tactic, and then decides which of the possible results to add to its solution.

One important difference between executing a tactic compared to directly executing its underlying
sequence of actions is that a tactic only exposes a single result: the result of its last action.
This scoping boundary helps constrain future actions by limiting how many arguments are available
for invoking them in the current solution.

\subsection{Inducing tactics from solutions}

To learn tactics, we take an inductive approach: we aim to extract useful
tactics by generalizing segments of previous solutions. Our goal is to find
tactics that would have simplified several of those solutions had those tactics
been available to the agent.

Suppose we have a set $\mathcal{S}$ of solutions.
In Peano, each solution can be seen as a straight-line program\footnote{More generally,
the dependency structure between the actions, which are implied by the parameters,
generates a directed acyclic graph, but for simplicity we assume actions are fully ordered.}
that executes actions until the domain verifier determines that the solution satisfies the goal.
To find candidate tactics, we first extract each contiguous subsequence $s_{i:j}$ of length
at least 2 from each $s \in \mathcal{S}$.
Then, we take all pairs of same-length
subsequences and compute the tactic that is the \emph{least general generalization} of each
pair. In our restricted tactic language, this is
a simple case of anti-unification: if the two sequences call different actions,
then there is no generalization available in our language\footnote{In a more powerful
tactic language, for instance with loops and conditionals, we could aim to
encode more complex patterns into tactics. These extensions are interesting directions
for future work, where several techniques for inductive program synthesis could be applied.};
if they call the same actions, then there's always a tactic that generalizes those sequences, and
we can compute the most specific argument structure of their generalization by
only introducing new parameters if strictly necessary\footnote{We can see
all sequences that invoke the same actions as forming a bounded lattice,
and this anti-unification procedure as computing the \emph{meet} of two sequences, which might introduce
new tactic parameters. $\top$ is a sequence where all arguments in the tactic's body come from parameters. This structure is analogous to the subsumption lattice of first-order predicate calculus \cite{plotkin1972automatic}}.
This yields one tactic that, had it been available, could have simplified
at least the two input sequences of actions.

The procedure above gives us a set of candidate tactics $\mathcal{T}$
that generalize solution spans from $\mathcal{S}$.
To determine which of these candidates are worth making into new actions,
we first compute the number of segments in $\mathcal{S}$ that $t_i \in \mathcal{T}$ generalizes,
denoted by $m(t_i, \mathcal{S})$.
Multiplying that by $|t_i| - 1$ gives us how many actions in $\mathcal{S}$ would $t_i$ have saved if we
replaced $|t_i|$ actions by a single invocation of $t_i$.
In other words, $m(t_i, \mathcal{S}) (|t_i| - 1)$ is a measure of how
much can $t_i$ \emph{compress} the solutions in $\mathcal{S}$.
We take that quantity and divide it by the number of parameters of $t_i$ to
compute the \emph{utility} of $t_i$: $$u(t_i) = \frac{m(t_i, \mathcal{S})(|t_i| - 1)}{p(t_i)} \enspace .$$
Essentially, we seek tactics that are the least general explanation of the most solution steps in $\mathcal{S}$.

During learning, tactic induction can be introduced by alternating between the standard learning loop
of ConPoLe --- where it attempts to solve problems and improve its policy --- with learning tactics from
existing solutions. At each round, we add discovered tactics with an utility above a threshold $U_{min}$
to the agent's action space.
Furthermore, before policy training,
we rewrite all of the agent's solutions found so far using
the new tactics, everywhere they apply.
In this way, the policy is always trained on solutions that
are irreducible given the tactics induced up to that moment.

\subsection{Result: learning with tactic induction}
Figure~\ref{fig:learning-curves} shows the success rate across iterations of the
ConPoLe agent trained with a tactic induction step after each batch of problems.
While ConPoLe alone fails to make progress in the later three sections, only
managing to solve degenerate equations (e.g., $x + 0 = 10$ or $1\times x + 0 = 5$),
tactic induction allows the agent to make progress and eventually solve all problems.

\begin{figure}
    \centering
    \includegraphics[width=\textwidth]{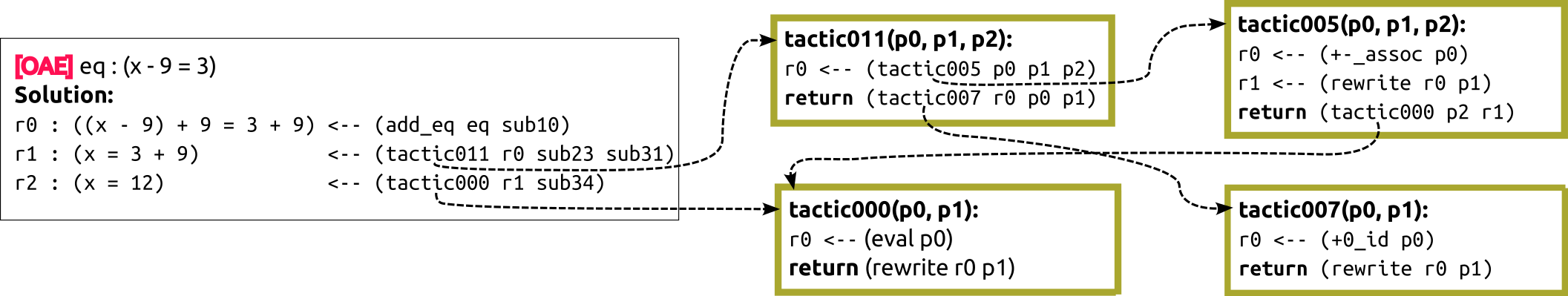}
    \caption{Solution found by the agent to a ``One-step Addition Equation''.
    In terms of the tactics at this moment of training, the solution can be expressed
    with 3 steps. These steps have a probability of $1.28 \times 10^{-6}$ to be
    generated by an agent taking random actions when the action space contains the axioms
    and the 18 tactics learned so far. In contrast, given just the axioms, the probability
    of finding the equivalent solution at random is $10^{-12}$.}
    \label{fig:solution-with-tactics}
\end{figure}

In addition to enabling the agent to solve the harder problems,
we find that the hierarchy of tactics induced during training
reflects how the Khan Academy sections conceptually build on each other.
Figure~\ref{fig:solution-with-tactics} shows the solution found to equation from the
One-Step Addition and Subtraction Equations section, the first section that ConPoLe alone does 
not manage to solve. Tactics constructed from previous solutions allows this
problem to be solved within 3 steps: first, applying the axiom introduced in this
section of adding a constant to both sides, and then applying two induced tactics that each
simplify one of the sides of the equation. The left-hand side has an expression involving
$x$, and thus requires ``combining like terms''; the right-hand side is a simple expression
with constants which can be fully evaluated.

After enough examples of equations like this are seen, the agent induces a tactic that
solves them in one step. In terms of that tactic, together with others learned
in the ``One-step Multiplication and Division Equation'', the ``Two-step Equations''
can indeed be solved with two steps, as the section name suggests.
However, that is only true when steps have the appropriate conceptual level.
Tactic induction enables to agent to construct those steps and successfully
exploit how exercises build on previously developed concepts.

\section{Curriculum Construction from Abstractions}
\label{sec:curriculum}

Our results suggest that our ability to construct abstraction is key in the
acquisition of mathematical knowledge: it's only possible to reason about
advanced results once we have developed the necessary abstractions.
But when learning mathematics, humans do not start from scratch: we follow
carefully constructed sequences of pedagogical experiences 
(i.e., curricula) developed by previous generations.
Following a well-designed curriculum has an unusual effectiveness in mathematics:
even discoveries that took many generations of bright individuals to emerge and develop,
such as calculus or complex analysis, later become part of traditional high-school
or college-level classes.
What makes a sensible curriculum for a learner, and why can this
structure speed-up learning so effectively?

We hypothesize that the abstractions underlying the domains of interest can shed light on these 
questions. Abstractions --- such as our tactics --- can be built by composition of simpler
abstractions.
This structure suggests a partial order $\prec_\mathcal{T}$ on tactics $t_i \in \mathcal{T}$
where $t_1 \prec t_2$ if $t_2$ depends on $t_1$.
If abstractions are \emph{induced}, i.e. extracted from concrete experiences (such as solutions),
then a teacher that wants to help a learner induce $t_2$ would
naturally place the learning
experiences leading to $t_1$ first. Moreover, clustering the learning experiences
that suggest an abstraction is necessary might catalyze that process.
Two natural questions related to this induced curriculum arise.
How well does it agree with the curriculum designed by human educators?
And how effective would that ordering be in accelerating the learning of
a second agent?

Our setup allows us to empirically explore these questions in the 
case study of formalized Khan Academy sections.
Let $\mathcal{P}$ be the set of training problems seen by an agent,
and suppose $\mathcal{T}$ is the set of all of the agent's induced tactics
at the end of training.
From $\mathcal{T}$, we can compute each tactic's \emph{dependency set} $DS(t_i)$ by
taking all tactics on which $t_i$ depends directly or indirectly;
this corresponds to all tactics
that precede $t_i$ in the transitive closure of $\prec_\mathcal{T}$.
This partial order on tactics lets us infer a corresponding partial order
$\prec_{\mathcal{P}}$ on problems by comparing the tactics that their solutions use.
More precisely, suppose $(p_a, T_a)$ and $(p_b, T_b)$ are two
problem/tactic set pairs, where $T_a$ (resp. $T_b$)
is the set of tactics invoked in the solution found for $p_a$ (resp. $p_b$).
We would like to decide which of these
problems should come first in a curriculum for a learner.
A natural choice is to consider that $p_a \prec_{\mathcal{P}} p_b$
whenever $p_a$ depends on strictly less tactics than $p_b$,
i.e. 

\[
p_a \prec_{\mathcal{P}} p_b\enspace \iff \enspace \bigcup_i DS(T_a^{(i)}) \subset \bigcup_i DS(T_b^{(i)}) \enspace .
\]

Given $\prec_{\mathcal{P}}$, any topological ordering of the problems $\mathcal{P}$
would yield a curriculum that is compatible with the structure of abstractions
induced from solving $\mathcal{P}$.
Since $\prec_{\mathcal{P}}$ is \emph{partial} --- two problems are incomparable if each uses tactics not present in the solution to the other ---, multiple orders
might agree with $\prec_{\mathcal{P}}$\footnote{It is also possible in principle
to find two solutions to the same problem that use different sets of tactics.
This is rare in our setup in practice, but more generally many consistent
ways to aggregate the sets of tactics associated to a problem would still
yield a valid partial order.}.
We first ask: how do these
orders compare to the human-designed order of Khan Academy sections?

To answer this question, we first sample multiple orderings of $\mathcal{P}$
by running a simple stochastic topological sorting algorithm that respects
$\prec_{\mathcal{P}}$ but otherwise chooses which element to put next at
random from the set of candidate next problems. Then, for each problem
in the resulting order, we take the index of the section of Khan Academy
where the problem originated from (1 to 5),
and compute the Kendall tau distance \cite{kendall1955rank} to the Khan Academy order
(which, in this case, is simply the number of inversions in the generated list
of integers).

\begin{figure}
    \centering
    \includegraphics[width=0.8\textwidth]{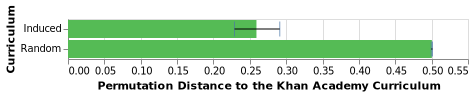}
    \caption{Comparison between the Khan Academy curriculum and the curriculum inferred from an agent
             with tactic induction trained on problems seen in a random order.
             Bars indicate the average number of inversions --- adjacent swaps needed to make the curricula agree ---
             with 99\% confidence intervals (averaged over a random sample of topological orderings).
             For comparison, we show the number of inversions that a random permutation of the problems
             produces.}
    \label{fig:inversions}
\end{figure}

Figure~\ref{fig:inversions} compares the curricula we obtain by
topological orderings that respect $\prec_{\mathcal{P}}$ with random orderings
of problems. In each case, we sample 100 curricula and compute the normalized Kendall tau distances to the Khan Academy ordering, and compute 99\% bootstrapped
confidence intervals. Curricula induced from $\prec_{\mathcal{P}}$ are
significantly closer to the Khan Academy ordering, indicating that the underlying abstractions capture part of what makes a curriculum pedagogically sensible.

\begin{figure}
    \centering
    \includegraphics[width=\textwidth]{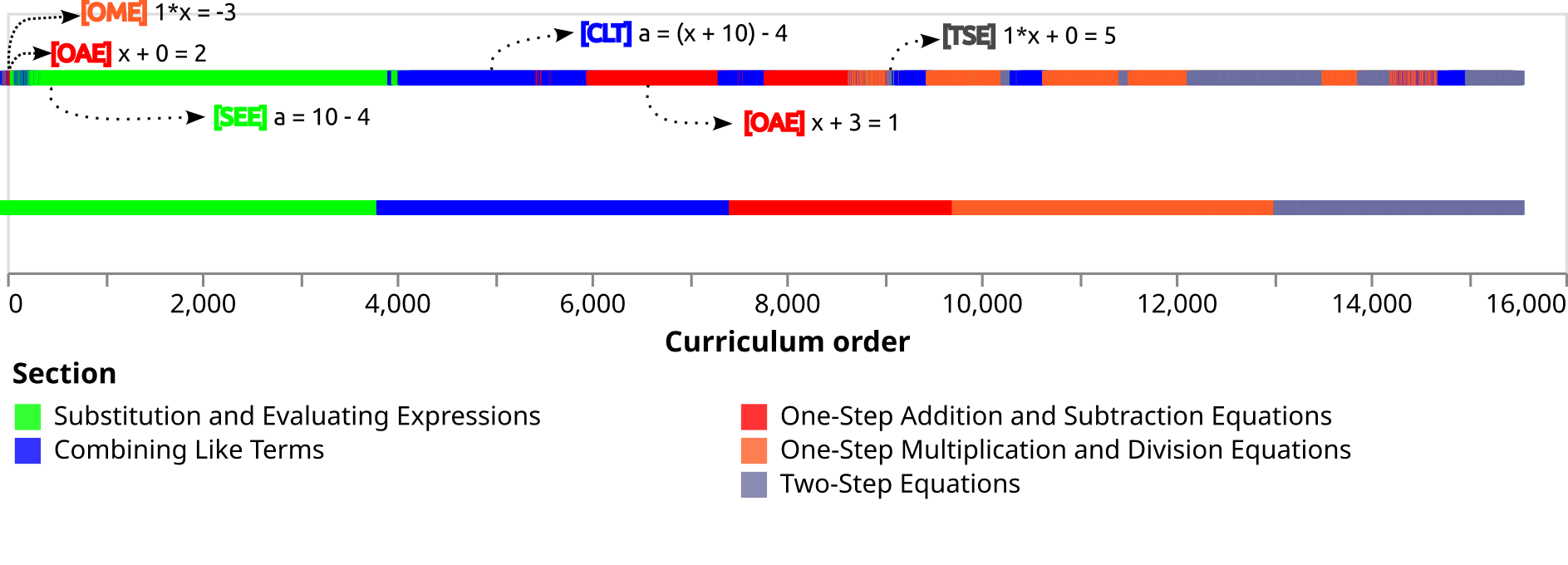}
    \caption{Sample induced curriculum from agent's learned abstractions.}
    \label{fig:reconstructed-curriculum}
\end{figure}

Figure~\ref{fig:reconstructed-curriculum} shows a sample induced curriculum,
comparing it to the order from Khan Academy.
The ordering of Khan Academy sections is partly recovered by the abstractions, though the
induced curricula have a much more fine-grained dependency structure.
For example, the abstractions alone suggest that the degenerate ``one-step equation''
$x + 0 = 2$ can be solved directly from the axioms.
After specific cases of combining like terms have been learned, the induced
curriculum already allows the equations depending on those cases to come.

Identifying fine conceptual dependencies at a problem level could be useful
for automated tutoring systems, which might use them to suggest problems or give worked examples to a particular student in a personalized manner.
On the other hand, Khan Academy sections have instructional content in addition to exercises.
Thus, their section structure also needs to take
into account which concepts are most easily taught together, a preference that our model does not have.

\subsection{Result: the synergy between tactic induction and curricula}

We now evaluate whether this induced curriculum would help a second agent learn faster.
Intuitively, when solving problems in a random order, the first agent takes many samples to accumulate
evidence that a certain abstraction is useful, and time is wasted attempting problems that are either
too hard (require abstractions several levels above its current tactics) or too easy (e.g., solved within
a single step given its current tactics).
Thus, the fact that the learner is performing tactic induction
provides a strong reason why a curriculum might be helpful.

To train a second agent using a curriculum, we first sample
one of the valid topological orders on all problems solved
by the first agent.
Then, we split the resulting sequence of problems into 3 blocks.
When sampling problems for the second agent, we initially
only sample problems from the first block, until training
success rate reaches a minimum of $90\%$, at which point we
include the second block in the pool of problems, and again
include the third block once the agent reaches a success
rate of $90\%$ when solving the last batch of problems.
This schema matches a well-known definition of a training curriculum
\cite{bengio2009curriculum}: a function that assigns
weights to training examples at each iteration in such a way
that, at the end of training, sampling from the reweighted distribution
is equivalent to sampling from the original target distribution,
but examples might be gradually introduced throughout training.

With this setup, we observe marked benefits from training using
a curriculum. In Figure~\ref{fig:learning-curves}, we observe that
agents trained on a curriculum induced from abstractions (green) indeed learn faster
than the agent that discovered the abstractions without a curriculum (blue) --- in particular,
the second agent learns to solve the last section significantly earlier. Even our simple curriculum scheme allows the agent
to learn useful abstractions much faster, avoiding problems
that are too hard (i.e., would require much higher abstractions
than those constructed so far).
We'd expect this effect to get even more pronounced if we were
formalizing a markedly larger domain, where only a small
fraction of all new problems could be reasonably solved
given the available tactics. As this fraction diminishes,
so does the likelihood of randomly selecting a problem from
which an agent like ours --- trained on its successful searches ---
can learn productively.


This result suggests a computational account of a cultural ratchet \cite{tomasello2009cultural} effect of mathematics: once one generation makes mathematical discoveries,
captured in a hierarchy of new abstractions
and relations between them, the next generation learns from a carefully constructed order
so that it can reach the same point of understanding much faster. In our case, our agents
learned until they reached ceiling performance in the domains we modelled.
But if each of our agents had a limited budget compared to how long it would take to fully learn the target domains --- much like humans have limited lifetimes ---
we would still observe an inter-generational speed-up. In this case, the first agent would
not reach mastery of all domains,
but would still be able to construct a curriculum from its experiences.
A second agent would learn to reach the same performance in much less time, and would be able to productively explore further in it's lifetime.

\section{Discussion and Conclusion}

We introduced Peano, a language for expressing mathematical reasoning
and an associated environment for solving problems formally in a
finite action space.
Using Peano, we formalized 5 sections of the algebra curriculum
from the Khan Academy educational platform.
Search alone is unable to find solutions to non-trivial problems
because of a combinatorial explosion of the search space.
Reinforcement learning provides a means to learn from past searches
and make progress, but even then, longer solutions are unreachable.
But combining reinforcement learning with abstraction learning
--- in the form of tactic induction, where we learn useful reusable
components from solutions found so far --- allows an agent to make progress, learning to solve all problems across our 5 sections of algebra.

In addition to enabling an agent to solve problems more effectively, we have found
abstraction learning to match our intuition about which higher-level
skills students need to master when learning basic algebra.
This was reflected in the fact that reordering problems using the dependencies between abstractions in their solutions largely recovers the Khan Academy curriculum ordering.
Note that human curricula are designed with more than problem ordering
in mind: they also aim to facilitate instruction, where conceptual cohesion is important.
These aspects are irrelevant for our agents, since they only learned from attempting problems.
Even so, the dependencies implied by the learned abstractions recovered some of the structure
present in the human-designed curriculum,
suggesting that they capture a key aspect of curriculum design.

Moreover, ordering problems based on abstractions interacts
favorably with our model of inductive learning of abstractions.
Such an ordering can focus the agent on problems that elicit new abstractions
which are concisely expressed in terms of the agent's already learned abstractions.
This focus allows the agent to quickly accumulate examples from which the
new abstractions can be induced, avoiding both \emph{too easy} or \emph{too hard}
problems. Indeed, we have observed this reconstructed curriculum to accelerate learning
of a ``second-generation'' agent.

Together, these results provide a computational account of the importance of
abstraction learning for human mathematics. On the one hand, tactic induction
dramatically helps an individual learner in leveraging experience gained in easier problems
to solve harder ones. On the other, after a certain domain has been mastered, the hierarchical
structure of the learner's induced abstractions allow it to structure the learning
of future generations, helping them arrive faster at the point the first learner left off.

We believe these experiments bring an important insight towards the goal of
having artificial agents to perform human-like mathematical reasoning.
Much of the past research in this direction has been devoted to
making search methods more effective through search heuristics
--- either hand-crafted, learned, or a combination of both.
But given any set of initial axioms, many problems will inevitably be out of reach of search\footnote{The analogy to program synthesis makes this point clear:
if the base programming language is x86 Assembly, synthesizing a simple
list-sorting algorithm would be an enormous challenge.
One would certainly not hope the synthesizer to stumble upon a correct
implementation of merge sort in the search space of sequences of x86 instructions.
But in a high-level domain-specific language with convenient operations
for list processing (e.g., a function to merge two sorted lists),
even merge sort becomes short and much less surprising.
This insight carries over to mathematical reasoning when we see solutions (or proofs)
as programs.}.
Abstraction learning provides a means to progress much beyond the limit of search methods.

Several avenues for future work naturally arise.
First, the Peano environment has limitations: it does not allow the prover to
create intermediate lambda terms (that is, sub-lemmas), and does not support backwards reasoning.
Both these capabilities would be needed to allow natural solutions to problems
arising in some educational domains we have not yet tackled.
For instance, proofs by induction essentially require the prover to produce
one lemma per inductive case,
which corresponds to a lambda abstraction (e.g., in the case of natural numbers,
the lambda would be a function that takes $n$ and a proof of the proposition
for $n$, and outputs a proof of the proposition for $n + 1$).
Similarly, in the case of induction, the first step is typically backwards:
when one realizes that one must prove a proposition for all natural numbers,
it is natural to start by claiming the proof will be by induction; then,
the necessary ``lemmas'' are proved.
One challenge is extending Peano to support these natural moves
while maintaining a finite action space.

Second, the tactics we were able to learn in Section~\ref{sec:tactic-induction}
are rather simple, consisting of short straight-line programs.
Our tactic language cannot express natural high-level
actions involving repetitions and conditions, such as
``apply commutativity and associativity until you group $x$ and $-x$''
or ``evaluate all operations you can''.
Extending our tactic language along with the tactic induction algorithm will
be necessary to extend our method to more complex domains.
Tactic induction becomes more challenging, but potentially more powerful,
as the environment itself becomes more expressive. Many techniques from
inductive program synthesis can be potentially helpful for that direction.

Third, our approach to library learning involves learning tactics, but not new theorems.
Tactics can give hints at useful auxiliary theorems to be proven.
For example, a tactic that produces proofs that $n^2 > 0$ when given several integers $n$
suggests that there is a general procedure for producing such proofs, and
that therefore $\forall n, n^2 > 0$ is a theorem.
As we move towards covering more complex mathematical domains, we believe this notion of library
learning --- of useful results in addition to solution strategies --- will be important.

Finally, our approach to learning tactics relies on first encountering many examples of its use:
a tactic is deemed useful if it would have been helpful in solving many past problems.
But human mathematician are often able to perform this inductive leap after
just a single example.
For an example, consider Erd\"os' lower bound on the Ramsey number $R(s)$;
see \cite{gowers2000two} for a discussion of this result and its significance.
This is a result to a purely combinatorial problem that applies a probabilistic
argument in a surprising way. This single example is most often enough
for human mathematicians to realize a potentially fruitful new tactic;
this has indeed turned into the Probabilistic Method, now widely used in combinatorics.
The fact that this tactic stands out so clearly when one reads about this result
poses several puzzling questions.
What makes such a solution so \emph{surprising},
and certain parts of it especially \emph{interesting}?
Answers for these questions would help us understand a core notion in the human
practice mathematics: not only some statements are true and some are false,
but some are more \emph{interesting} than others.
Computationally characterizing what \emph{interestingness} means
might be an important goal towards having computers be able to provide insights
into human mathematics \cite{colton2000notion}.
After all, that would imply not just proving new results,
but also recognizing which ones are significant.

\dataccess{All the code, data and configuration files needed to reproduce the experiments in this paper are available online at \url{https://github.com/gpoesia/peano}}
\funding{This work was supported by a NSF Expeditions Grant, Award Number (FAIN) 1918771, and by the Stanford HAI Hoffman--Yee project ``AI Tutors to Help Prepare Students for the 21st Century Workforce''. GP was also supported by a Stanford Interdiciplinary Graduate Fellowship (SIGF).}


\bibliographystyle{plain}

\bibliography{references.bib}











\end{document}